\documentclass[lettersize,journal]{IEEEtran}

\usepackage[belowskip=-10pt,aboveskip=3pt]{caption}
\usepackage{textcomp}
\usepackage{xcolor}
\usepackage{graphicx}
\usepackage{hyperref}

\hypersetup{
    colorlinks,
    linkcolor={blue!50!black},
    citecolor={blue!50!black},
    urlcolor={blue!80!black}
}

\def\BibTeX{{\rm B\kern-.05em{\sc i\kern-.025em b}\kern-.08em
    T\kern-.1667em\lower.7ex\hbox{E}\kern-.125emX}}
    
\usepackage{geometry}\usepackage{amsmath}
\usepackage{amssymb}
\usepackage{algorithm}
\usepackage{rotating}
\usepackage{algpseudocode}
\usepackage{acronym}
\usepackage{multirow}
\usepackage{siunitx}
\usepackage{graphicx}
\usepackage{tikz}
\usetikzlibrary{matrix,shapes,arrows,positioning,chains}

\usepackage{makecell}

\usepackage{acronym} 
\usepackage{hyperref}
\usepackage[export]{adjustbox}
\usepackage{tabularx}
\usepackage{caption}
\newcommand{\orcid}[1]{\href{https://orcid.org/#1}{\includegraphics[width=8pt]{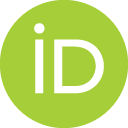}}}

\DeclareMathOperator*{\argmin}{argmin}
\begin{document}

\setlength{\abovedisplayskip}{2pt}
\setlength{\belowdisplayskip}{2pt}

\author{\IEEEauthorblockN{Shirwan Piroti \orcid{0009-0007-8954-2293}\,, 
Ashima Chawla \orcid{0000-0001-5933-3107}\,, 
Tahar Zanouda \orcid{0009-0005-3646-600X}}\

%\
%\IEEEauthorblockA{\IEEEauthorrefmark{1}Ericsson, Engineering
%Unit Integrated RAN, Kista, Sweden}

%\IEEEauthorblockA{\IEEEauthorrefmark{2}Ericsson, Network
%Management Lab, Athlone, Ireland
%  }
% \IEEEauthorblockA{\IEEEauthorrefmark{3}Global AI Accelerator,
%Ericsson AB, Kista, Sweden
%   }
\IEEEauthorblockA{Ericsson.\\
\texttt{\{shirwan.piroti, ashima.chawla, tahar.zanouda}\}@ericsson.com}}

\title{Mobile Network Configuration Recommendation using Deep Generative Graph Neural Network}

% \title{Configuration Management Framework For
% Telecom Networks using Deep Generative Graph
% Neural Network}

\maketitle

% add page numbers 
%\thispagestyle{plain}
%\pagestyle{plain}

\begin{abstract}
%Radio Access Networks (RAN) have numerous configurable parameters. making configuration labor-intensive due to RAN complexity, deployment diversity, and varying hardware technologies. Traditional methods rely on domain knowledge for individual parameter configuration, often leading to sub-optimal results. To improve this, a framework using a Deep Generative Graph Neural Network (GNN) is proposed. It encodes the network into a graph, extracts subgraphs for each RAN node, and employs a Siamese GNN (S-GNN) to learn embeddings. The framework recommends configuration parameters for a multitude of parameters and detects misconfigurations, handling both network expansion and existing cell reconfiguration. Tested on real-world data, the model surpasses baselines, demonstrating accuracy, generalizability, and robustness against concept drift.

There are vast number of configurable parameters in a Radio Access Telecom Network. A significant amount of these parameters is configured by Radio Node or cell based on their deployment setting. Traditional methods rely on domain knowledge for individual parameter configuration, often leading to sub-optimal results. To improve this, a framework using a Deep Generative Graph Neural Network (GNN) is proposed. It encodes the network into a graph, extracts subgraphs for each RAN node, and employs a Siamese GNN (S-GNN) to learn embeddings. The framework recommends configuration parameters for a multitude of parameters and detects misconfigurations, handling both network expansion and existing cell reconfiguration. Tested on real-world data, the model surpasses baselines, demonstrating accuracy, generalizability, and robustness against concept drift.

\end{abstract}

\section{ Introduction}
Accurate configuration in a Radio Access Network (RAN) is crucial for stable and reliable connectivity, ensuring Key Performance Indicators (KPIs) are within acceptable ranges. RAN consists of interconnected Radio Nodes, each with unique parameters for specific area coverage. Hardware and software selection depends on the mobile operator and service area requirements, and are set up with Configuration Management (CM) parameters \cite{CM3GPP}. RAN Performance is monitored via Performance Management (PM) counters.

The configuration process in RAN is complex, often involving vendor default settings and additional adjustments based on deployment scenarios. Deployment typically includes two stages: Node Provisioning and Post-integration, with the latter requiring reconfigurations to achieve KPI acceptance. To address these challenges, this paper introduces a Deep Generative GNN, which aims to learn optimal parameters from stable networks and recommend configurations for other network areas. The approach involves three stages:

\begin{itemize}
    \item \textbf{Stage I:} Graph-based representation of multi-dimensional RAN data.
    \item \textbf{Stage II:} Creation of subgraphs using a graph-sampling method and training a Siamese Graph Neural Network (S-GNN) with data from stable networks.
    \item \textbf{Stage III:} Recommendation of Configuration Values tailored for different scenarios:
    \begin{itemize}
        \item Expansion: Installing new cells on existing Radio Nodes.
        \item Greenfield Deployment: Setting up new Radio Nodes in regions without existing infrastructure.
        \item Modification: Adjusting existing cells and Radio Nodes to correct misconfigurations.
    \end{itemize}
\end{itemize}

The method aims to develop a machine-assisted method for recommending network configuration parameters, transferring knowledge from stable operational networks, and identifying telecom system misconfigurations.

The key contributions of this work include:
\begin{enumerate}
    \item Developing an efficient configuration system with minimal manual intervention.
    \item Representing multidimensional RAN data as vectors in an embedding space.
    \item Introducing a robust, generalizable one-shot learning solution \cite{oneshotLearner} that automatically updates with new data.
\end{enumerate}

The paper's structure is as follows: Section II introduces the methodology, and Section III  details the evaluation metrics. Section IV details the results. Finally, section V concludes the paper with future research directions.

\section{Methodology}
This section details the model architectures and the method for generating configuration values. %It introduces a framework for configuration management, designed to learn optimal parameters from stable networks and recommend configurations for other network areas. 
The approach involves three stages, as shown in Figure 1.

% The method focuses on creating a machine-assisted system for suggesting network configuration parameters, transferring knowledge from operational networks, and identifying misconfigurations in telecom systems. It gathers RAN data from existing mobile networks and employs a graph neural network generative model to understand the hierarchical nature of this data. By inputting subgraphs and producing corresponding embeddings, the model ensures that subgraphs with similar contexts and configurations are closely aligned in the embedding space. This facilitates recommending appropriate configuration parameters for similar network scenarios.

% \begin{enumerate}
%     \item Utilizing RAN topology and configuration data to derive consistent configuration parameters for new Radio Nodes through the generative graph neural network model.
%     \item Configuring various node attributes concurrently in telecom networks.
% \end{enumerate}

\subsection{Data Description}\label{sec:Pre-processing}
The data for this study is collected from multiple eNBs and gNBs. This section details how data was collected and processed.

\subsubsection{Feature Values and Pre-processing}

For this end, each cell in the telecom network is represented by two feature vectors in the dataset:
\begin{itemize}
    \item \( x \) - Predictor values defined during cell planning, including LTE attributes  \cite{CM3GPP} like Channelbandwidth (data transmission capacity) and Earfcndl (central downlink frequency channel number), and NR attributes such as bSChannelBwDL and arfcnDL  \cite{CM3GPP}.
    
    \item \( y \) - Configuration values displaying network variation, encompassing LTE attributes  \cite{CM3GPP} like pZeroNominalPusch (target power level received by the eNB per resource block) and preambleInitialReceivedTargetPower (initial preamble power value), and NR attributes such as endcUlNrLowQualThresh and rachPreambleRecTargetPower  \cite{CM3GPP}.
\end{itemize}

\begin{figure}[htp!]
    \centering
    \includegraphics[width=\linewidth]{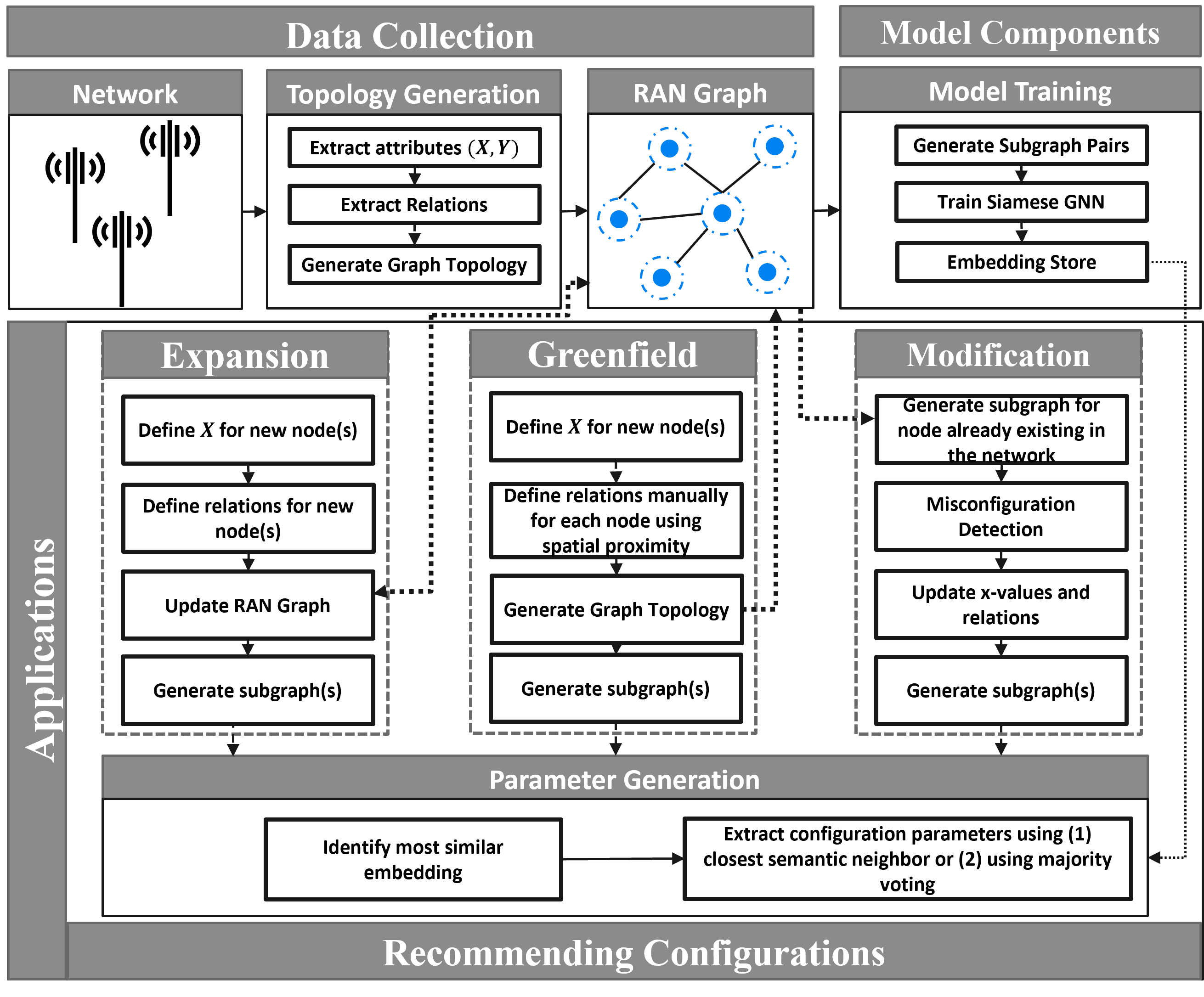}
    \caption{Overview of model training and configuration generation for diverse use cases.}
    \label{fig:recommend_config}
\end{figure}

These vectors combine LTE and NR attributes, facilitating model generalization across different technology deployments (4G, 5G, mixed mode), where \textit{N} and \textit{M} denote the number of LTE and NR cells in the network, respectively, is defined:
\begin{align*}
    (X_{LTE},Y_{LTE})=\{x_{LTE,i},y_{LTE,i} \}_{i=1}^N, \\
    (X_{NR},Y_{NR})=\{x_{NR,i},y_{NR,i} \}_{i=1}^M,
\end{align*}

The datasets are then preprocessed as described in the following steps:
\begin{itemize}
    \item  normalize values in each set to the range \textit{[0,1]}
    \item  Concatenate sets:
    \begin{align*}
        (X,Y)=(X_{LTE}||X_{NR},Y_{LTE}||Y_{NR})
    \end{align*}
    
%Concatenation is the union of the LTE- and NR-sets. The set of features for LTE and NR can be different so concatenation of the sets introduce missing values such that each $x\in X$ and $y \in Y$ have the same dimension, respectively. The missing values appear where a feature is available for LTE but not for NR, and vice versa.
    \item Impute missing values by concatenating with 0.
\end{itemize}

\subsection{RAN Graph}

The RAN is modeled as an undirected graph $G = (V, E)$ with vertices $V=\{V_i\}_{i=1}^{N+M}$, each representing a RAN cell, and edges $E$ defining cell and inter-Radio Node relations, where \begin{equation*}
    E \subseteq \big\{V_i,V_j:(V_i,V_j)\in V^2, i \not = j \forall 0\leq i,j \leq N+M \big\}.
\end{equation*}

Subgraphs are $G_i = (V_i \cup \tilde{V}_i, E_i)$, with $\tilde{V}_i$ being a random uniform sample of neighbors from $N(V_i)$, the vertices connected to $V_i$. This approach focuses on local neighborhoods for configuring Radio Nodes or cells.

%The RAN is represented as an undirected graph: $G = (V,E)$, where $V=\{V_i\}_{i=1}^{N+M}$ is the set of vertices in the graph where each vertex corresponds to a cell in the RAN and
%\begin{equation*}
%    E \subseteq \big\{(V_i,V_j:(V_i,V_j)\in V^2, i \not = j \forall 0\leq i,j \leq N+M \big\}
%\end{equation*}
%
%is the set of relations between cells in the network. These  are composed of cell relations and inter-Radio Node relations.

%A subgraph is defined as $G_i = (V_i \bigcup \Tilde{V}_i, E_i)$, where
%\begin{equation*}
%    \Tilde{V}_i \overset{U}{\leftarrow} N(V_i) s.t. \Tilde{V}_i \subseteq N(V_i)
%\end{equation*}

%$N(V_i)$ denotes the set of vertices which $V_i$ shares an edge with and $\Tilde{V_i}$ is the set of neighbors that are randomly sampled uniformly without replacement from $N(V_i)$. Subgraphs are leveraged as it is assumed that only the local neighborhood is relevant when configuring Radio Nodes or cells. The cardinality of $N(V_i)$ is bounded by the number of cells in the network which can result in larger subgraphs than needed to make accurate predictions.

\subsection{Datasets for inductive Graph Neural Networks}
GNNs are a class of machine learning models designed for learning and reasoning over graph-structured data. Unlike transductive GNNs, inductive GNNs \cite{TransGNN} can generalize to unseen nodes and graphs. They achieve this by leveraging shared weights and aggregating information from neighboring nodes to learn node representations. Each subgraph $G_i$ has an associated feature set which is defined as:

\begin{equation}
    \Tilde{X}_i=\{x_j\}_{V_j\in V_i\bigcup\Tilde{V}_i}.
\end{equation}
Also, $y_i$ is the vector containing configuration values for the center cell $V_j$ for which the subgraph is constructed around. The dataset is then defined as:
\begin{equation}
    \mathcal{D}^{(N+M)}=\{(G_i,\Tilde{X}_i,y_i)\}_{i=1}^{N+M}
\end{equation}

\subsection{Model Definition \& Architecture}
\begin{figure}[htp!]
    \centering
    \includegraphics[width=\linewidth]{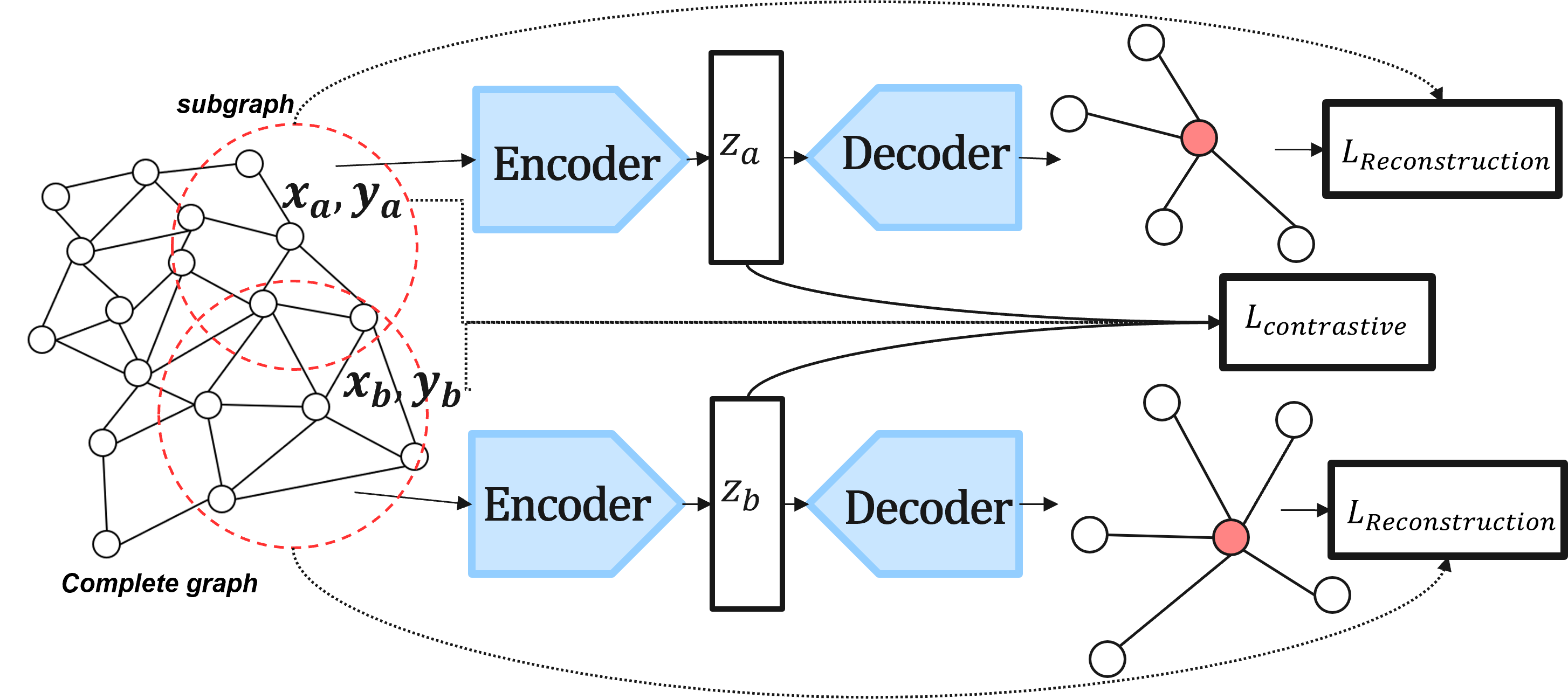}
    \caption{End-to-end model architecture for configuration management framework.}
    \label{fig:E2Earchitecture}
\end{figure}

The models implemented in this study are one-shot meta-learners. The main model is an S-GNN-based model \cite{SGNN} that is trained using self-supervision through minimization of a contrastive loss. Figure \ref{fig:E2Earchitecture} depicts the architecture of the model. The S-GNN is compared to a Graph Auto-Encoder (GAE) model that is trained through unsupervised learning.

The S-GNN and the GAE are functions represented by GNNs that are parameterized by $\omega$ and are defined as:

\begin{align}
    f_{S-GNN_\omega}:(G_i,\Tilde{X}_i)\rightarrow (G_i,Z_i), \\
    f_{GAE_\omega}=f_{Decoder_\omega} \circ f_{Encoder_\omega} \\
    f_{Encoder_\omega}:(G_i,\Tilde{X}_i)\rightarrow(G_i,Z_i), \\
    f_{Decoder_\omega}:(G_i,Z_i)\rightarrow(G_i,\hat{X}_i).
\end{align}

$f_{Encoder_\omega}$ is the encoder part of the GAE and generates the latent space representation and $f_{Decoder_\omega}$ reconstructs the input from the latent space. $Z_i$ is the set of all embeddings of vertices associated with subgraph $G_i$, where \textit{d} is a hyperparameter that refers to the dimension of the latent space. 
\begin{equation}
    Z_i=\{z_j\}_{V_j \in V_i\cup \Tilde{V}_i} and \:  z_j\in \mathbb{R}^d
\end{equation}

\subsection{Architecture \& Training}

The neural networks $f_{s-GNN_{\omega}}$ , $f_{Encoder_{\omega}}$, $f_{Decoder_{\omega}}$, each compose multi-headed Graph Attention (GATv2Conv) \cite{GATv2Conv} and aggregates the output of each head using a Feed Forward Network (FFN). 

% The forward propagation through the networks is detailed in equations:

% \begin{equation}
%     \~{h}_{o,i} = \alpha_{i, j} \omega_{o} h_{i} + \sum _{V_{j} \in \~V_{i}} \alpha_{i, j} \omega_{o} h_{i}
% \end{equation}

% $\alpha$ are the attention weights and are defined as:

% \begin{equation}
%  \alpha_{i,j}=  \frac{{exp(\alpha_{o}^{T}LeakyReLU(\omega_{o}[h_{i}||h_{j}}])}{{\sum _{V_{j} \in V_{i} \bigcup \~{V}_{i} }exp(\alpha_{o}^{T}LeakyReLU(\omega_{o}[h_{i}||h_{j}}])}
% \end{equation}
    
% where $\alpha_{o}$ and $\omega_{0}$ are vectors containing learnable parameters.

% \begin{equation}
%   h^{'}_{i}=  \overset O {\underset{o=1}\parallel} h^{'}_{o,i}
% \end{equation}

% is the output of the GATv2Conv operator which is then fed to a $l-layer MLP f_{MLP} $ to combine the outputs from the individual heads, yielding

% \begin{equation}
%     h_{i}^{''} = f_{MLP} (h_{i}^{'}): \mathbb{R}^{dim(h_{i}^{'}) \rightarrow \mathbb{R} \frac{dim(h_{i}^{'}}{O}}
% \end{equation}

% Where $O$ denotes the number of heads in the GAT. Figure 2 illustrates the integrated end-to-end model architecture, which combines both S-GNN and GAE components.

% \subsection{Training}
The parameters $\omega$ in $f_{S-GNN_{\omega}}$ are trained with the objective to minimize the loss  with $M$ as a hyperparameter:

\begin{multline}
\mathcal{L}_{contrastive}(c_{a,b}, z_a, z_b) = (1+c_{a,b})||z_{a} - z_{b}||_{2}  \\
+ (1- c_{a,b}) max(0,M) - ||z_{a}- z_{b}||_{2}.
\end{multline}

\begin{equation}
    c_{a,b} = 2  \frac {y_{a} \cdot y_{b}} { (||y_{a}||_{2}||y_{b}||_{2}) }-1 \\
\end{equation}

$c_{a,b}$ are the labels, indicating how similar two arbitrary cells $V_{a}$ and $V_{b}$ are based on their corresponding configuration setups. Additionally, Subspace Learning is leveraged to limit the number of training pairs by identifying the most informative samples \cite{SubspaceLearning}. Informative pairs are composed of ambiguous samples, that is, samples that are close in the embedding space but belong to different classes.

The parameters $\omega$ in $f_{GAE_{\omega}}$ are trained with the objective to minimize the loss:
\begin{equation}
\mathcal{L}_{reconstruction} = \frac{1}{|\tilde{X}_{n|}} \sum^{|\tilde{X}_{n}|}_{j=1} ||x_{j} - \tilde{x_{j}}||_{2}
\end{equation}

However, only $f_{Encoder_{\omega}}$ will be used to infer the configuration values.

\subsection{Inference}
The function that yields the configuration values for the $k:th$ cell that is introduced to the network after training is defined as:
\begin{equation}
 y_{N+M+k}   = f((G_{N+M+k}, \tilde{X_{N+M+k}}) | f_{\omega, D^{(N+M+k-1)}})
\end{equation}

% \begin{equation}
%     \hat{y}^{(0)}_{N+M+k}=\\
%     f((G_{N+M+k},\Tilde{X}_{N+M+k})|f_\omega,\mathcal{D}^{(N+M+k-1)})
% \end{equation}

where $f_\omega$ is either $f_{S-GNN_\omega}$ or $f_{Encoder_\omega}$, $\mathcal{D}^{(N+M+k-1)}$ is the dataset that includes the N+M samples used for training and the $k-1$ samples that were introduced after the initial training.

\begin{equation}
    \mathcal{F}_{N+M+k}^{(0)}=\{||z_{N+M+k}-z_i||_2\}_{i=1}^{N+M+k-1}=\{l_i\}_{i=1}^{N+M+k-1}
\end{equation}

Defines the set of all distances in the latent space from the new cell $V_{N+M+k} $, to all other existing cells. The configuration values are generated using one of two methods.

\subsubsection{Using closest semantic subgraph neighbor}
The configurations are acquired from the closest semantic subgraph in the embedding space.

% Defines the set of all distances in the latent space from the new cell $V_{N+M+k}$ to all other existing cells. The configuration values are generated using one of two methods.

    \begin{equation}
        \hat{y}_{N+M+k}^{(0)}= \argmin_{y_m \sim(G_m, \Tilde{X}_m, y_m)} \mathcal{F}^{(0)}_{N+M+k}
    \end{equation}

\subsubsection{Using Majority voting}
The configurations is acquired from an aggregation of the node's $ K $  closest neighbors in the latent space, where $ K $ is a pre-defined parameter of neighbors.

\begin{equation}
    \mathcal{F}^{(n+1)}_{N+M+k} = \mathcal{F}^{(n)}_{N+M+k} \backslash \{l^{(n)}_{n+M+k}\}
\end{equation}

$ l^{(n+1)}_{N+M+k} $ is the $n:th$ smallest value in $F^{0}_{N+M+k}$ and $\mathcal{F}^{n+1}_{N+M+k} $ is the set of loss values, excluding the set of losses: 
\begin{equation*}
    \{l^{(n)}\}^{n-1}_{i=1}
\end{equation*}
This yields the configurations, as shown in equation (\ref{configs}).
\begin{equation}
    \hat{y}_{N+M+k}^{(n)}=\argmin_{y_m\sim(G_m,\Tilde{X}_m,y_m)}\mathcal{F}_{N+M+k}^{(n)}
    \label{configs}
\end{equation}
where $m$ is an index value of the $y_m$ that yields
%\begin{equation*}
%    m\in[0,|\mathcal{L}_{N+M+k}|]\cap\mathbb{Z}_0^+
%\end{equation*}
$ \hat{y}_{N+M+k}^{(n)}$ which is the $n$:th most optimal set of configuration values. Thus,
\begin{equation}
     \hat{y}_{N+M+k}^{Majority}=aggregate\big(\{\hat{y}_{N+M+k}^{(n)}\}_{n=0}^{K-1}\big)
\end{equation}
where \textit{aggregate} is to be specified per attribute included in $y$.

\section{Evaluation Metrics}
This section details metrics used to evaluate the performance of the models. 
\begin{figure}[htp!]
    \centering
    \includegraphics[width=\linewidth]{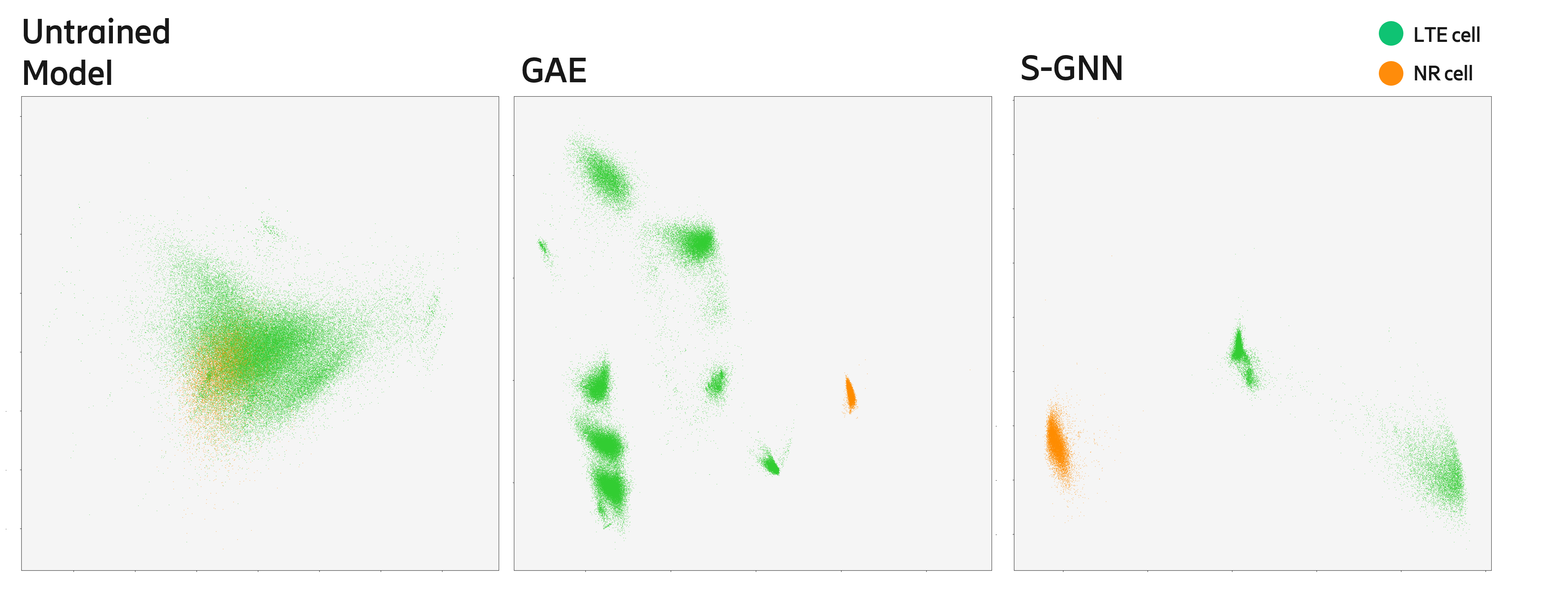}
    \caption{2-dimensional representation of the embeddings produced by three distinct models. Dimension of embeddings are redcued using \textit{PCA}.}
    \captionsetup{belowskip=0pt}
    \label{fig:embeddings}
\end{figure}

\subsection{Configuration Accuracy Score}

The accuracy of the generated configurations by the models across the whole network is then defined as:

\begin{equation}
    Accuracy = \frac{1}{N+M} \sum ^{N+M}_{j = 1} \frac {y_{j} \cdot \hat{y}_{j}} { ||y_{j}||_{2}||\hat{y}_{j}||_{2} }
    \label{accuracy}
\end{equation}

% \begin{equation}
%     cosineSimilarity( y_{j}, \hat{y}_{j}) = \frac {y_{j} \cdot \hat{y}_{j}} { (||y_{j}||_{2}||\hat{y}_{j}||_{2}) }
%     \label{cosine}
% \end{equation}

Given $\hat {y}_{j} = y_{m}$ are the configuration values for the $m:th$ cell network that yields the smallest $L_{2} - loss $ with the embedding of cell $ (G_{j}, \tilde{X_{j}})$.

\subsection{Misconfiguration Detection Score}
Anomaly detection is applied to evaluate the novelty of input samples in the network, providing operators with confidence values for the generated configurations. These values are anomaly scores, calculated by a model trained to identify novelties based on the distribution of embeddings in a latent space. If this score surpasses a predefined threshold, it signals the operator to reevaluate the specific instance.
\begin{equation}
    anomalyScore_{i} = f_{anomaly(z_{i} \sim f_{\omega} (G_{i}, \tilde{X_{i}})}).
\end{equation}
Figure \ref{fig:network} presents the embeddings' distributions with anomaly scores from iForest \cite{iforest}, indicating variation in network configurations.
\section{Results}

The paper evaluates two architectures, S-GNN and GAE, across four datasets. Their performance, measured using Average Optimal Cosine Similarity (\ref{accuracy}), is detailed in Table \ref{table}. Higher scores in this measure correlate with better configuration prediction in telecom networks. Figure \ref{fig:embeddings} illustrates the models' 14-dimensional embeddings reduced to 2 dimensions via Principal Component Analysis (PCA), comparing an untrained model, GAE, and S-GNN. This visual representation highlights the distinct learning patterns of each model. GAE, an unsupervised model, demonstrates high accuracy in discerning different deployment contexts in the RAN, despite not being trained on configuration data. This indicates a strong correlation between the contextual information it processes and the resulting configurations. 

In contrast, S-GNN shows higher accuracy in recognizing configuration setups when trained on both configuration and contextual data. However, this reliance on configuration data might lead to decreased adaptability over time as network settings evolve.

%The accuracy of the two models on 4 distinct datasets is presented in Table \ref{table}. The accuracy measure is the estimated Average Optimal Cosine Similarity \ref{accuracy}. A higher score indicates better ability to predict correct configurations.  Figure 3 depict a low dimensional representation of the embedding space. Each embedding produced by the neural networks is a 14-dimensionals vector. these are mapped to a 2-dimensional space using Principal Component Analysis (PCA) \cite{PCA}. Each image depicts the embeddings of 3 distinct models: an untrained model, GAE and S-GNN. The GAE is an unsupervised model and is able to identify different deployment contexts across the RAN. The high accuracy of the GAE indicates that the that the contextual information provided to the model is strongly correlated with the configurations, considering that GAE is not trained using configuration data. The S-GNN is able to identify different modes of configuration setups across the RAN.The accuracy of the S-GNN on the different dataset indicate that the model can recommend configurations with greater accuracy if configuration data is used for training together with contextual information. However, this make it more prone to performance degradation than the GAE as configurations in the RAN change over time.

\begin{figure}[htp!]
    \centering
    \includegraphics[width=\linewidth]{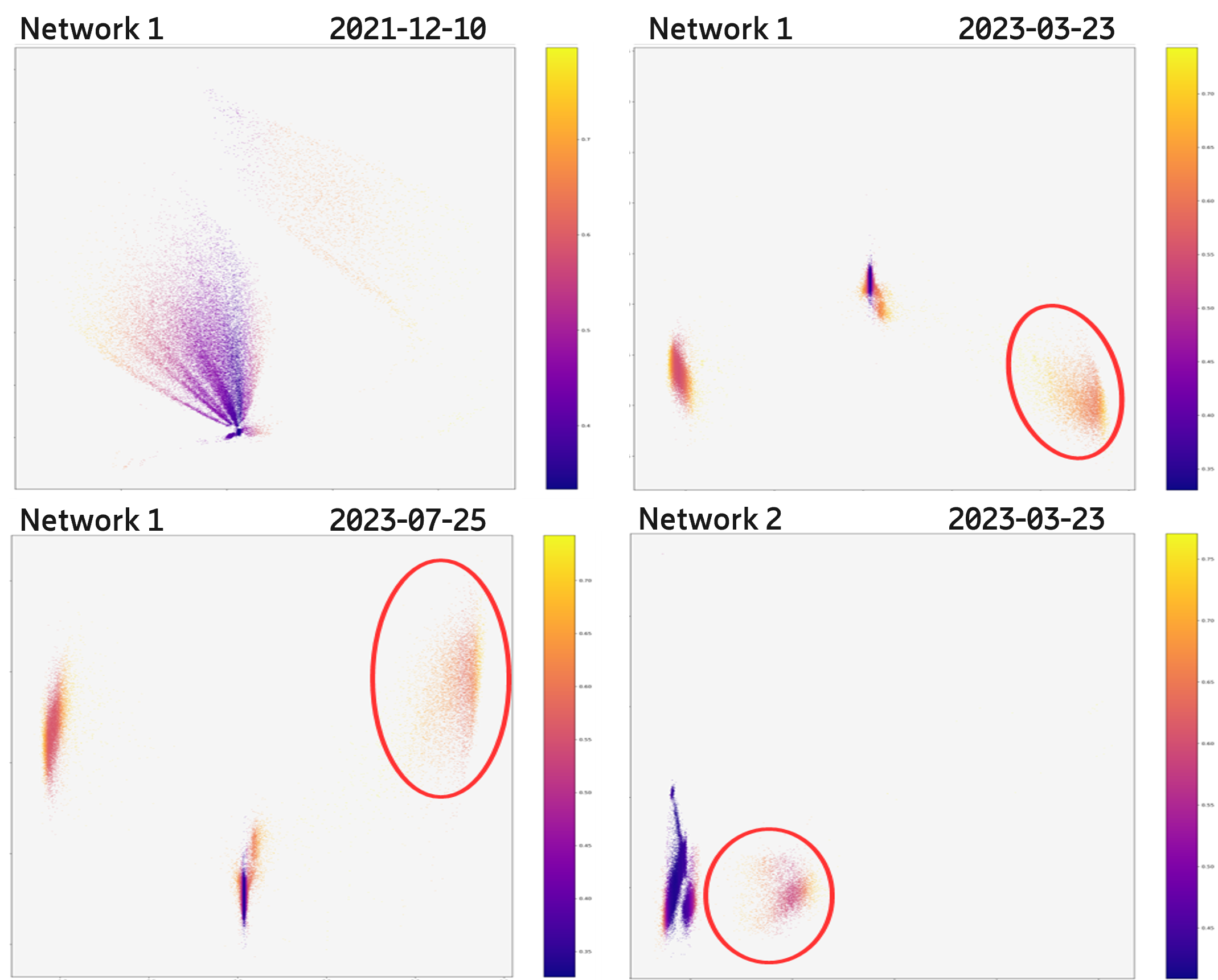}
    \caption{S-GNN's 2D embeddings represent novelty, with highlighted circles (embedding) for highest anomaly scores.}
    \label{fig:network}
\end{figure}

Figure \ref{fig:network} presents the embeddings' distributions with anomaly scores. Notably, networks with less configuration variation show higher model accuracy, suggesting the models, especially S-GNN, perform better in more stable configuration environments. In summary, while GAE excels in unsupervised learning from non-configuration data, S-GNN is more accurate in supervised scenarios but may face challenges in adapting network configurations changes.

%Figure 4 depict the distributions of the embeddings where the color is associated with an anomaly score. The anomaly scores are generated using iForest \cite{IsolationForest}. A greater value indicates a higher degree of novelty. The distribution of anomaly scores are also an indicator of degree of variation of configurations of across the RAN. Figure \ref{fig:network} shows that there is more variation in network 1. This also explain why the model yields higher accuracy in network 2. With less variation in configurations, the model is able to recommend configurations with greater accuracy.

\begin{table}[ht!]
\centering
\begin{tabular}{|l|c|c|c|r|}
\hline
\textbf{Model} & \multicolumn{3}{c|}{\textbf{Dataset}} & \textbf{Accuracy} \\
\cline{2-4}
 & \textbf{Type} & \textbf{Network} & \textbf{Date} & \\
\hline
S-GNN & Test & 1 & 2021-12-10 & 0.888 \\
S-GNN & Train & 1 & 2023-03-23 & \textbf{0.953} \\
S-GNN & Test & 1 & 2023-07-25 & \textbf{0.921} \\
S-GNN & Test & 2 & 2023-03-23 & 0.990 \\
\hline
GAE & Test & 1 & 2021-12-10 & \textbf{0.906} \\
GAE & Train & 1 & 2023-03-23 & 0.916 \\
GAE & Test & 1 & 2023-07-25 & 0.907 \\
GAE & Test & 2 & 2023-03-23 & \textbf{0.991} \\
\hline
\end{tabular}
\caption{Accuracy for the
GAE and the S-GNN, on 4 distinct datasets. The range is bounded in the range [0, 1].}
\label{table}
\end{table}

%All experiments were conducted using PyTorch 2.0.1 and Torch Geometric 2.3.1, with GAE model tuning via Optuna \cite{Optuna}. 

\newpage
\section{Conclusion and Future Direction}
In this paper, the proposed framework enables mobile operators to generate a multitude of values for configuration parameters in telecom networks. At the core of this framework, a Deep Generative Model combines GNN and Siamese Neural Network. The framework represents RAN as a graph and then leverages subgraphs to enable inductive inference. The model generates embeddings which are used to infer configurations and to detect misconfiguration in the network. The model was evaluated on multiple datasets from operational networks. The model yields high accuracy on the datasets, indicating that the framework is generalizable and robust against concept drift.

In future developments, this work will focus on challenges related to recommending configurations for O-RAN \cite{ORAN} networks. Additionally, future work will involve eXplainable Artificial Intelligence (XAI) to demystify AI decisions, enhancing trust and transparency between machines and users.

% \begin{enumerate}
%     \item \textbf{Network Planning Model:} Graph dataset edges signify RAN cell relations, established manually or via ANR post-deployment [17]. Pre-deployment requires manual definition of cell relations for this method.
    
%     \item \textbf{PM Data Integration:} The model is predicated on being trained with data from a stable network, and aims to factor in network performance in future configuration recommendations.
    
%     \item \textbf{Explainable Misconfiguration:} Future work will involve eXplainable Artificial Intelligence (XAI) [19] to demystify AI decisions, enhancing trust and transparency between machines and users.
% \end{enumerate}

\bibliographystyle{ieeetr}
\bibliography{bibliography}

\begin{thebibliography}{1}

\bibitem{CM3GPP}
3GPP, ``3gpp ts 32.300: Telecommunication management; configuration management (cm); name convention for managed objects,'' 2024.

\bibitem{oneshotLearner}
L.~Bertinetto, J.~F. Henriques, J.~Valmadre, P.~Torr, and A.~Vedaldi, ``Learning feed-forward one-shot learners,'' {\em Advances in neural information processing systems}, vol.~29, 2016.

\bibitem{TransGNN}
L.~Anghinoni, Y.-t. Zhu, D.~Ji, and L.~Zhao, ``Transgnn: A transductive graph neural network with graph dynamic embedding,'' in {\em 2023 International Joint Conference on Neural Networks (IJCNN)}, pp.~1--8, IEEE, 2023.

\bibitem{SGNN}
N.~Mehrotra, N.~Agarwal, P.~Gupta, S.~Anand, D.~Lo, and R.~Purandare, ``Modeling functional similarity in source code with graph-based siamese networks,'' {\em IEEE Transactions on Software Engineering}, vol.~48, no.~10, pp.~3771--3789, 2021.

\bibitem{GATv2Conv}
S.~Brody, U.~Alon, and E.~Yahav, ``How attentive are graph attention networks?,'' {\em arXiv preprint arXiv:2105.14491}, 2021.

\bibitem{SubspaceLearning}
E.~Krivosheev, M.~Atzeni, K.~Mirylenka, P.~Scotton, and F.~Casati, ``Siamese graph neural networks for data integration,'' {\em arXiv preprint arXiv:2001.06543}, 2020.

\bibitem{iforest}
X.~Zhao, Y.~Wu, D.~L. Lee, and W.~Cui, ``iforest: Interpreting random forests via visual analytics,'' {\em IEEE transactions on visualization and computer graphics}, 2018.

\bibitem{ORAN}
M.~Polese, L.~Bonati, S.~D’oro, S.~Basagni, and T.~Melodia, ``Understanding o-ran: Architecture, interfaces, algorithms, security, and research challenges,'' {\em IEEE Communications Surveys \& Tutorials}, 2023.

\end{thebibliography}

\vspace{12pt}
\end{document}